\documentclass[10pt,twocolumn,letterpaper]{article}

\usepackage{cvpr}              

\usepackage{graphicx}
\usepackage{amsmath}
\usepackage{amssymb}
\usepackage{booktabs}

\usepackage{color}
\usepackage{arydshln}
\usepackage{bm}
\usepackage{array}
\usepackage{comment}
\usepackage{caption}
\usepackage{colortbl} 
\usepackage{multirow}
\usepackage{makecell}
\usepackage{wrapfig}
\usepackage{paralist}
\usepackage{rotating}

\usepackage[ruled,vlined]{algorithm2e}
\usepackage[normalem]{ulem}

\usepackage[pagebackref,breaklinks,colorlinks]{hyperref}

\usepackage[capitalize]{cleveref}
\crefname{section}{Sec.}{Secs.}
\Crefname{section}{Section}{Sections}
\Crefname{table}{Table}{Tables}
\crefname{table}{Tab.}{Tabs.}

\def\cvprPaperID{} 
\def\confName{CVPR}
\def\confYear{2024}

\begin{document}


\title{Dysen-VDM: Empowering Dynamics-aware Text-to-Video Diffusion with LLMs}

\author{
Hao Fei\textsuperscript{1}\quad 
Shengqiong Wu\textsuperscript{1}\quad
Wei Ji\textsuperscript{1,}\thanks{Corresponding Author.}\quad
Hanwang Zhang\textsuperscript{2,3}\quad
Tat-Seng Chua\textsuperscript{1}\\
\textsuperscript{1}National University of Singapore \quad \textsuperscript{2}Skywork AI, Singapore \quad \textsuperscript{3}Nanyang Technological University\\
{\tt \{haofei37,swu,jiwei,dcscts\}@nus.edu.sg, hanwangzhang@ntu.edu.sg}
}

\maketitle

\begin{abstract}
Text-to-video (T2V) synthesis has gained increasing attention in the community, in which the recently emerged diffusion models (DMs) have promisingly shown stronger performance than the past approaches.
While existing state-of-the-art DMs are competent to achieve high-resolution video generation, they may largely suffer from key limitations (e.g., action occurrence disorders, crude video motions) with respect to the \emph{intricate temporal dynamics modeling}, one of the crux of video synthesis.
In this work, we investigate strengthening the awareness of video dynamics for DMs, for high-quality T2V generation.
Inspired by human intuition, we design an innovative dynamic scene manager (dubbed as \textbf{Dysen}) module, which includes 
(\textbf{step-1}) extracting from input text the key actions with proper time-order arrangement,
(\textbf{step-2}) transforming the action schedules into the dynamic scene graph (DSG) representations,
and (\textbf{step-3}) enriching the scenes in the DSG with sufficient and reasonable details.
Taking advantage of the existing powerful LLMs (e.g., ChatGPT) via in-context learning, Dysen realizes (nearly) human-level temporal dynamics understanding.
Finally, the resulting video DSG with rich action scene details is encoded as fine-grained spatio-temporal features, integrated into the backbone T2V DM for video generating.
Experiments on popular T2V datasets suggest that our Dysen-VDM consistently outperforms prior arts with significant margins, especially in scenarios with complex actions.
Codes at {\url{http://haofei.vip/Dysen-VDM/}}.
\end{abstract}

\section{Introduction}
\label{Introduction}

\vspace{-1mm}
Recently, AI-Generated Content (AIGC) has witnessed thrilling advancements and remarkable progress, e.g., ChatGPT~\cite{ouyang2022training}, DELLE-2~\cite{ramesh2022hierarchical} and Stable Diffusion (SD)~\cite{rombach2022high}. 
As one of the generative topics, text-to-video synthesis that generates video content complying with the provided textual description has received an increasing number of attention in the community.
Prior researches develop a variety of methods for T2V, including generative adversarial networks (GANs) \cite{abs-1810-02419,SaitoSKK20,MunozZAB21}, variational autoencoders (VAEs) \cite{li2018video,abs-2104-10157,DuanLWYT22}, flow-based models \cite{BashiriWLJMDDTS21,KumarBEFLDK20}, and auto-regressive models (ARMs) \cite{KalchbrennerOSD17,WeissenbornTU20,GeHYYPJHP22}.
More recently, diffusion models (DMs) have emerged to provide a new paradigm of T2V.
Compared with previous models, DMs advance in superior generation quality and scaling capability to large datasets \cite{HarveyNMWW22,abs-2206-07696}, and thus showing great potential on this track \cite{abs-2212-00235,abs-2303-08320,abs-2302-07685,abs-2303-13744}.

\begin{figure}[!t]
\begin{center}
\includegraphics[width=0.87\linewidth]{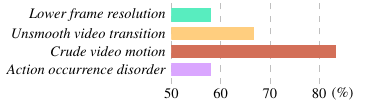}
\end{center}
\vspace{-1.2em}
\caption{
Common issues in the existing text-to-video (T2V) synthesis.
We run the video diffusion model (VDM) \cite{HoSGC0F22} with random 100 prompts, and ask users to summarize the problems.
}
\label{intro}
\vspace{-4mm}
\end{figure}

Although achieving the current state-of-the-art (SoTA) generative performance, DM-based T2V still faces several common yet non-negligible challenges.
As summarized in Figure \ref{intro}, four typical issues can be found in a diffusion-based T2V model, such as \emph{lower frame resolution}, \emph{unsmooth video transition}, \emph{crude video motion} and \emph{action occurrence disorder}.
While the latest DM-based T2V explorations paid much effort into enhancing the quality of video frames, i.e., generating high-resolution images \cite{abs-2211-11018,abs-2302-07685,abs-2303-13744}, they may largely overlook the modeling of the \textbf{intricate video temporal dynamics}, the real crux of high-quality video synthesis, i.e., for relieving the last three types of aforementioned issues.
According to our observation, the key bottleneck is rooted in the nature of video-text modality heterogeneity:
language can describe complex actions with few succinct and abstract words (e.g., predicates and modifiers), 
whereas video requires specific and often redundant frames to render an action.

\begin{figure*}[!t]
\centering
\includegraphics[width=0.99\textwidth]{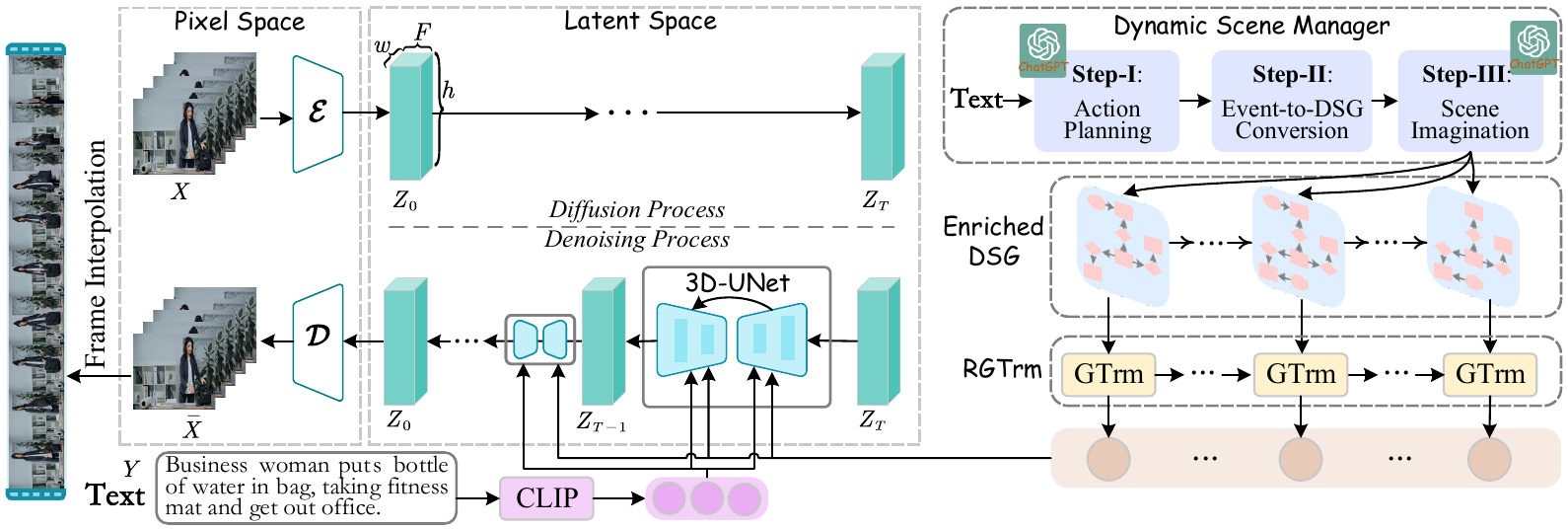}
\caption{
The architecture of the dynamics-aware T2V diffusion model, Dysen-VDM.
The dynamic scene manager (Dysen) module 
operates over the input text prompt and produces the enriched dynamic scene graph (DSG), which is encoded into the resulting fine-grained spatio-temporal scene features are integrated into the video generation (denoising) process.
}
\label{framework}
\vspace{-4mm}
\end{figure*}

Picturing that, whenever we humans create a film from a given instruction, we always first extract the key actions from the instruction into an event playlist with time order.
We then enrich the simple events with more possible specific scenes, i.e., with our imagination.
With such integral \emph{screenplay}, it can be effortless to project the whole video successfully.
Correspondingly, from the above intuition we can draw four key points of effective T2V modeling, especially for the scenario with complex dynamics.
\textbf{First}, sequential language mentions a set of movements that may not necessarily coincide with the physical order of occurrence, it is thus pivotal to properly organize the semantic chronological order of events.
\textbf{Second}, as prompt texts would not cover all action scenes, reasonable enrichment of video scenes is indispensable to produce delicate videos with detailed movements.
\textbf{Third}, the above processes should be carried out based on effective representations of structured semantics, to maintain the imagination of high-controllable dynamic scenes.
\textbf{Finally}, fine-grained spatio-temporal features modeling should be realized for temporally coherent video generation.

Based on the above observations, in this work we present a nichetargeting solution to achieve high-quality T2V generation by strengthening the awareness of video dynamics.
We propose a dynamics-aware T2V diffusion model, as shown in Figure \ref{framework}, in which we first employ the existing SoTA video DM (VDM) as the backbone T2V synthesis, and meanwhile devise an innovative \underline{\bf dy}namic \underline{\bf s}c\underline{\bf en}e manager (namely \textbf{Dysen}) module for video dynamics modeling.
To realize the human-level temporal dynamics understanding of video, we take advantage of the current most powerful LLM, e.g., OpenAI ChatGPT (GPT3.5/GPT4); we treat ChatGPT as the consultant for action planning and scene imagination in Dysen.
Specifically, in \textbf{step-I}, we extract the key actions from the input text, which are properly arranged in physically occurring orders.
In \textbf{step-II}, we then convert these ordered actions into sequential dynamic scene graph (DSG) representations \cite{JiK0N20}.
DSGs represent the intrinsic spatial\&temporal characteristic of videos in semantic structures, allowing effective and controllable video scene management \cite{LiYX22}.
In \textbf{step-III}, we enrich the scenes in the DSG with sufficient and reasonable details.
We elicit the knowledge from ChatGPT with the in-context learning \cite{abs-2201-11903}.
At last, the resulting DSGs with well-enriched scene details are encoded with a novel recurrent graph Transformer, where the learned delicate fine-grained spatio-temporal features are integrated into the backbone T2V DM for generating high-quality fluent video.

We evaluate our framework on the popular T2V datasets, including UCF-101 \cite{abs-1212-0402}, MSR-VTT \cite{XuMYR16}, as well as the action-complex ActivityNet \cite{KrishnaHRFN17}, where our model consistently outperforms existing SoTA methods on both the automatic and human evaluations with significant margins.
We show that our Dysen-VDM system can generate videos in higher motion faithfulness, richer dynamic scenes, and more fluent video transitions, and especially improves on the scenarios with complicated actions.
Further in-depth analyses are shown for a better understanding of how each part of our methods advances.

Overall, this paper addresses the crux of high-quality T2V synthesis by strengthening the motion dynamics modeling in diffusion models.
We contribute in multiple aspects.
\textbf{\em (i)} To our knowledge, this is the first attempt to leverage the LLMs for action planning and scene imagination, realizing the human-level temporal dynamics understanding for T2V generation.
\textbf{\em (ii)} We enhance the dynamic scene controllability in diffusion-based T2V synthesis with the guidance of dynamic scene graph representations.
\textbf{\em (iii)} Our system empirically pushes the current arts of T2V synthesis on benchmark datasets.
Our codes will be open later to facilitate the community.

\vspace{-1mm}
\section{Related Work}
\label{Related Work}

\vspace{-1mm}
Synthesizing videos from given textual instructions, i.e., T2V, has long been one of the key topics in generative AI.
A sequence of prior works has proposed different generative neural models for T2V.
Initially, many attempts extend the GANs \cite{GoodfellowPMXWOCB14} models from image generation \cite{abs-2008-05865,XuZHZGH018,ZhuP0019} to video generation \cite{clark2019adversarial,GordonP20,KahembweR20,FoxTET21,SkorokhodovTE22}.
While GANs often suffer from the issue of mode collapse leading to hard scalability, other approaches have proposed learning the distribution with better mode coverage and video quality than GAN-based approaches, such as VAEs \cite{li2018video,abs-2104-10157,DuanLWYT22}, flow-based models \cite{BashiriWLJMDDTS21,KumarBEFLDK20} and ARMs \cite{KalchbrennerOSD17,WeissenbornTU20,GeHYYPJHP22}.
Recently, diffusion models \cite{ho2020denoising} have emerged, which learn a gradual iterative denoising process from the Gaussian distribution to the data distribution, generating high-quality samples with wide mode coverage.
Diffusion-based T2V methods help bring better results with more stable training \cite{HarveyNMWW22,abs-2210-02303,abs-2206-07696,abs-2212-00235,abs-2303-08320,qu2023layoutllm,wu2024imagine}.
Further, latent diffusion models (LDMs) \cite{RombachBLEO22} have been proposed to learn the data distribution from low-dimensional latent space, which helps sufficiently reduce the computation costs, and thus receive increasing attention for T2V synthesis \cite{he2022latent,abs-2211-11018,abs-2302-07685,abs-2303-13744}.
In this work, we inherit the advance of LDMs, and adopt it as our backbone T2V synthesizer.

Compared with the text-to-image (T2I) generation \cite{XuZHZGH018,ZhuP0019,SongME21,RombachBLEO22} that mainly focuses on producing static visions in high-fidelity resolutions, T2V further places the emphasis on the modeling both of spatial\&temporal semantics, especially the scene dynamics.
Previously, some T2V research explores video dynamics modeling for generating high-quality videos \cite{VondrickPT16,YushchenkoA019,YuTMKK0S22,luo2023decomposed,esser2023structure}, i.e., higher temporal fluency, and complex motions, while they may largely be limited to the coarse-level operations, such as the spatio-temporal convolutions \cite{VondrickPT16}.
In the line of DM-based T2V \cite{zhang2023show,xing2023simda,ge2023preserve,lin2023videodirectorgpt,wang2023lavie,zhao2023motiondirector,wu2023tune,wang2023videocomposer}, most of the methods consider improving the video quality by enhancing the frame resolution \cite{abs-2211-11018,abs-2302-07685,abs-2303-13744}, instead of the perception of dynamics.
Most of the LDM-based T2V work also uses the spatio-temporal factorized convolutions in the 3D-UNet decoder \cite{he2022latent,abs-2211-11018,wang2023videofactory,wang2023modelscope}.
For example, \cite{abs-2304-08477} tries to strengthen motion awareness with a temporal shift operation.
All of these attempts, unfortunately, can be seen as a type of coarse-grained modeling.
In this work, we take fine-grained spatio-temporal feature modeling based on DSG representations.
We propose a systematic solution to enhance the diffusion awareness of the action dynamics.

\section{Preliminary}
\label{Task Modeling}

\vspace{-1mm}
\subsection{Text-to-video Latent Diffusion Model}
\label{Text-to-video Latent Diffusion Model}

\vspace{-1mm}
We first formalize T2V task as generating an video $X$=$\{x_1,\cdots, x_F\} \in \mathbb{R}^{F \times H \times W \times C}$ that specifies the desired content in the input prompt text $Y$=$\{w_1,\cdots,w_S\}$.
Here $F, H, W, C$ are the frame length, height, width, and channel number of video, respectively.
A latent diffusion model (LDM) is adopted for T2V which performs a forward (diffusion) process and a reverse (denoising) process in the video latent space.
Firstly, an encoder $\mathcal{E}$ maps the video frames into the lower-dimension latent space, i.e., $Z_0 = \mathcal{E}(X)$, and later a decoder $\mathcal{D}$ re-maps the latent variable to the video, $X = \mathcal{D}(Z_0)$.
Given the compressed latent code $Z_0$, LDM gradually corrupts it into a pure Gaussian noise $Z_T \sim \mathcal{N}(Z_T, 0, I)$ over $T$ steps by increasingly adding noisy, formulated as $q(Z_{1:T} |Z_0) = \prod_{t=1}^{T}q(Z_t|Z_{t-1})$.
and the learned reverse process $ p_{\theta}(Z_{0:T}) = p(Z_T)\prod_{t=1}^{T} p_{\theta} (Z_{t-1}|Z_t, Y)$ gradually reduces the noise towards the data distribution conditioned on the text $Y$.
T2V LDM is trained on video-text pairs ($X, Y$) to gradually estimate the noise $\epsilon$ added to the latent code given a noisy latent $Z_t$, timestep $t$, and conditioning text $Y$:
\begin{equation}\small\label{LDM}
\setlength\abovedisplayskip{4pt}
\setlength\belowdisplayskip{4pt}
 \mathcal{L}_{\text{\scriptsize LDM}} = \mathbb{E}_{Z \sim \mathcal{E}(X),Y,\epsilon,t} \left[ \, || \epsilon -\epsilon_{\theta} (Z_t, t, \mathcal{C}(Y))   ||^2 \, \right]  \,,
\end{equation}
where $\mathcal{C}(Y)$ denotes a text encoder that models the conditional text, and the denoising network $\varepsilon_\theta (\cdot)$ is often implemented via a 3D-UNet \cite{HoSGC0F22}, as illustrated in Figure \ref{framework}.

\vspace{-1mm}
\subsection{Dynamic Scene Graph Representation}
\label{Dynamic Scene Graph Representation}

\vspace{-1mm}
DSG \cite{JiK0N20} is a list of single visual SG of each video frame, organized in time-sequential order.
We denote an DSG as $G$=$\{G_1,\cdots,G_M\}$, with each SG ($G_m$) corresponding to the frame ($x_m$).
An SG contains three types of nodes, i.e., \emph{object}, \emph{attribute}, and \emph{relation}, in which some scene objects are connected in certain relations, forming the spatially semantic triplets `\emph{subject-predicate-object}'.
Also, objects are directly linked with the attribute nodes as the modifiers.
Besides, since a video comes with inherent continuity of actions, the SG structure in DSG is always temporal-consistent across frames.
This characterizes DSGs with spatial\&temporal modeling.
Figure \ref{framework} (right part) simply visualizes a DSG.

\begin{figure*}[!t]
\centering
\includegraphics[width=0.99\textwidth]{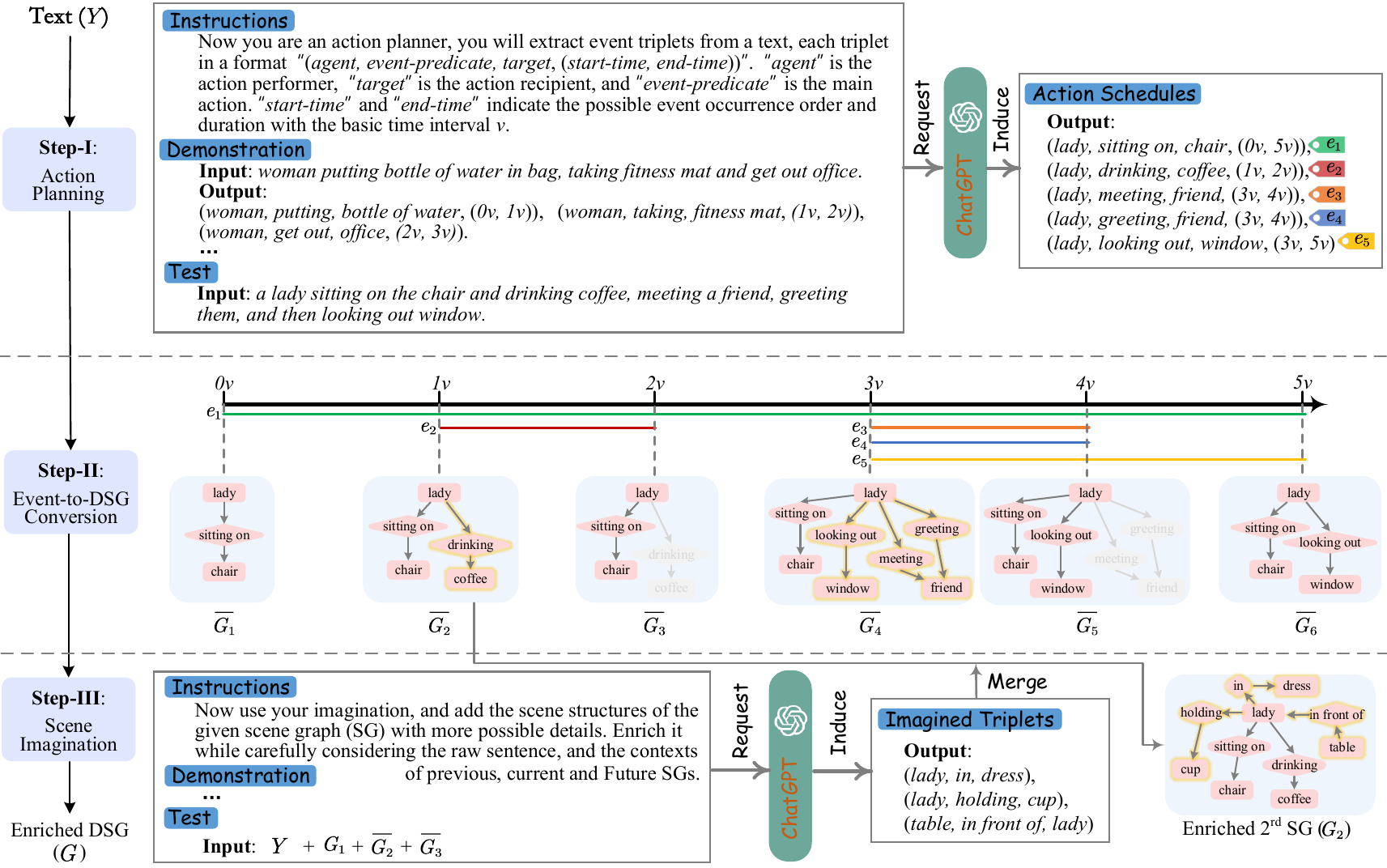}
\vspace{-1mm}
\caption{
Based on the given text, Dysen module carries out three steps of operations to obtain enriched DSG: 1) action planning, 2) event-to-DSG conversion, and 3) scene imagination,
where we take advantage of ChatGPT (i.e., GPT3.5 or GPT4) with in-context learning.
}
\label{DSM}
\vspace{-4mm}
\end{figure*}

\section{Methodology}
\label{Methodology}

\vspace{-1mm}
\textbf{Overall Framework.}
The architecture of our proposed dynamics-aware T2V diffusion framework is shown in Figure \ref{framework}.
The backbone T2V synthesizer is an LDM (cf. $\S$\ref{Text-to-video Latent Diffusion Model}).
During the denoising, the dynamic scene manager (Dysen) module (cf. $\S$\ref{Dynamic Scene Manager}) effectively captures the intrinsic spatial-temporal characteristic of input texts to better guide the T2V generation in the mainstay LDM (cf. $\S$\ref{Scene Integration for T2V Generation}).

\vspace{-1mm}
\subsection{Dynamic Scene Manager}
\label{Dynamic Scene Manager}

\vspace{-1mm}
As cast earlier, the input language instruction can often be succinct and abstract, which causes trouble generating concrete dynamic visual scenes in videos, especially when the actions are semantically complex.
To bridge the gap of dynamic scenes between texts and videos, here we propose a dynamic scene manager.
With Dysen, we carry out three steps of operations: action planning, event-to-DSG conversion, and scene enrichment.
Recently, the rise of LLMs has revealed the amazing potentials \cite{chowdhery2022palm,abs-2303-08774,touvron2023llama}, among which, ChatGPT \cite{ouyang2022training} is the most outstanding one in content understanding,
and perceiving complex events from language and comprehending the dynamic scenes in the way humans do \cite{abs-2304-03893,wu2023visual}.
Thus, we elicit such action planning and scene imagination abilities from ChatGPT.

\vspace{-4mm}
\paragraph{Step-I: Action Planning.}
We first ask ChatGPT to extract the key actions from the prompt texts.
Technically, we employ in-context learning (ICL) \cite{abs-2201-11903}.
We write the prompts, which include 1) a job description (\texttt{Instruction}), 2) a few input-output in-context examples (\texttt{Demonstration}), and 3) the desired testing text (\texttt{Test}).
Feeding the ICL prompts, we expect ChatGPT to return the desired action plans, as illustrated in Figure \ref{DSM}.
Specifically, we represent an action scene as ``(\emph{agent, event-predicate, target}, (\emph{start-time, end-time}))'',
in which `\emph{agent, event-predicate, target}' is the event triplet corresponding to the relational triplets as described in DSG (cf. $\S$\ref{Dynamic Scene Graph Representation});
`\emph{start-time, end-time}' is the temporal interval of this event.
Note that the atomic time interval is assumed as $v$, which is disentangled from a physical time duration.
Both the event scene triplets and the time arrangements are decided via ChatGPT's understanding of the input.
This way, even complex actions with multiple overlapped or concurrent events will be well supported.

\vspace{-4mm}
\paragraph{Step-II: Event-to-DSG Conversion.}
With the event schedule at hand, we then transform it into a holistic DSG structure.
Note that this DSG can be quite primitive, as each SG structure within DSG almost contains one triplet, which can be seen as the skeleton of the dynamic scenes.
Specifically, we construct the DSG along with the time axis incrementally, i.e., with each frame step having a corresponding SG.
According to the occurrence order and duration of events, in each frame we \emph{add} or \emph{remove} a triplet, until handling the last event.
This also ensures the SG at each frame step is globally unique.
The resulting DSG well represents the skeleton spatial-temporal feature of the events behind the input.
Figure \ref{DSM} illustrates the conversion process.

\vspace{-4mm}
\paragraph{Step-III: Scene Imagination.}
Based on the above initial DSG (denoted as $\overline{G}$=$\{\overline{G}_1,\cdots,\overline{G}_M\}$), we finally enrich the scenes within each SG.
For example, for each SG, there should be visually abundant scenes, e.g., objects will have various possible attributes, and different objects can be correlated with new feasible relations within the scene.
Also, the temporal changes between SG frames should be reflected, instead of the constant SG across a period.
This is intuitively important because all the events are continuous in time, and all the motions happen smoothly, e.g., in `\emph{a person sitting on a chair}', the motion `\emph{sitting}' can be broken down into a consecutive motion chain: `\emph{approaching}'$\to$`\emph{near to}'$\to$`\emph{sitting}'.
We again adopt the ChatGPT to complete the job, as it is effective in offering rich and reasonable imagination \cite{guo2023can,guo2023close}.
Concretely, the scene enrichment has two rounds.
The first round preliminarily enriches each SG one by one, i.e., by either \emph{adding} some new triplets or \emph{changing} the existing triplets.
As shown in Figure \ref{DSM}, the ICL technique is again used to prompt ChatGPT to yield the triplets to be added for the current SG, given the raw input text.
To ensure the dynamic scene coherency, we consider a \emph{sliding window context} (SWC) mechanism, when operating for the current SG, takes into account the current, previous (enriched), and following contexts of SGs (e.g., [$G_{m-1}$, $\overline{G}_m$, $\overline{G}_{m+1}$]).
This way, $\overline{G}_m$ can inherit from the previous well-established scene $G_{m-1}$, and meanwhile decide what to add or change to better transit to the next scene $\overline{G}_{m+1}$.
The second round further reviews and polishes the overall scenes of DSG from a global viewpoint, also via ChatGPT in another ICL prompting process.
This ensures all the actions go more reasonably and consistently.
The resulting final DSG is denoted as $G$=$\{G_1,\cdots,G_M\}$).

\vspace{-1mm}
\subsection{Scene Integration for T2V Generation}
\label{Scene Integration for T2V Generation}

\vspace{-1mm}
The enriched DSG ($G$) entails fine-grained spatial and temporal features.
Instead of using the general graph neural networks to encode the DSG structure, e.g., GCN, GAT, and RGNN \cite{marcheggiani-titov-2017-encoding,VelickovicCCRLB18,NicolicioiuDL19}, we consider the Transformer architecture \cite{VaswaniSPUJGKP17} that allows highly-parallel computation with the self-attention calculation.
To further model the temporal dynamics of the graphs, we consider the recurrent graph Transformer (RGTrm).
RGTrm has $L$ stacked layers, with a total of $M$ recurrent steps of propagation for each SG.
The representation $H^{l}_m$ of SG $G_m$ of $l$-th layer is updated as:
\setlength\abovedisplayskip{4pt}
\setlength\belowdisplayskip{4pt}
\begin{gather}\label{RGTrm}
H^{l+1}_m = O^l_k  \, ||_{k=1} \, ( 
\sum_{j \in \mathcal{N}_i} w_{i,j,m}^{k,l} \, V^{k,l}_m
) \,, \\
w_{i,j,m}^{k,l} = \text{Softmax}_j (
\frac{ \hat{Q}^{k,l}_m \cdot K^{k,l}_m }{ \sqrt{d_k} }
) \cdot E^{k,l}_m  \,, \\
\hat{Q}^{k,l}_m = (1-z_m) \cdot Q^{k,l}_{m-1} + z_m \cdot Q^{k,l}_m \,,  \\
z_m = \sigma ({W}^z  \cdot Q^{k,l}_m \cdot K^{k,l}_m ) \,,
\end{gather}
where 
$k$ denotes the attention head number.
$O^l_k$ is the $k$-th attention head representation.
$K^{k,l}_m$=${W}^{K}H^{l}_m$, $Q^{k,l}_m$=${W}^{Q}H^{l}_m$, $V^{k,l}_m$=${W}^{V}H^{l}_m$ are the key, query and value representations in the Transformer.
$E^{k,l}_m$=${W}^{E}\{e_{i,j,m}\}$ is the embedding of edge $e_{i,j,m}$ in DSG.
And $||$ is the concatenation operation.
We denote the final DSG representation as $H^{G}=\{H^{G}_1,\cdots,H^G_M\}$, where $H^{G}_m$ is the one of $m$-th SG ($G_m$).

\begin{figure}[!th]
\begin{center}
\includegraphics[width=0.98\linewidth]{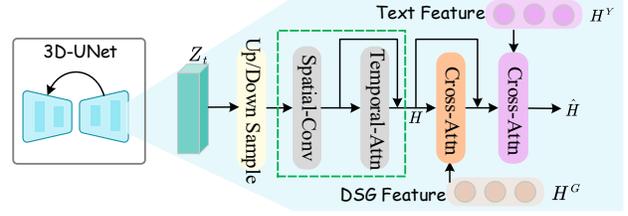}
\end{center}
\vspace{-1em}
\caption{
Illustration of the DSG integration.
}
\label{fig:scene_guidance}
\vspace{-4mm}
\end{figure}

\vspace{-2pt}
Next, we integrate the fine-grained spatial-temporal DSG features ($H^{G}$) into the 3D-UNet decoder for enhanced T2V generation, i.e., denoising process.
Although the original 3D-UNet \cite{HoSGC0F22} has a spatial-temporal feature modeling (as the green dotted box in Figure \ref{fig:scene_guidance}), it is limited by the coarse-grained operations, e.g., convolutions over 3D patches and attention over frames.
Thus, we insert an additional Transformer block with cross-attention for fusing the fine-grained $H^{G}$ representations, followed by another cross-attention to further fuse the raw text feature $H^{Y}$:
\begin{equation}\small
\setlength\abovedisplayskip{4pt}
\setlength\belowdisplayskip{4pt}
    \hat{H} = \text{Softmax}(\frac{H \cdot H^{G}} {\sqrt{d}}) \cdot H^{G} \,, \quad
    \hat{H} \gets \text{Softmax}(\frac{\hat{H} \cdot H^{Y}} {\sqrt{d}}) \cdot H^{Y}  \,,
\end{equation}
where $H$ is the coarse-grained spatio-temporal features.
The text representation $H^{Y}$ is encoded by CLIP \cite{RadfordKHRGASAM21}.

\subsection{Overall Training}

The overall training of Dysen-VDM system entails three major steps.

\begin{compactitem}

\item \textbf{Stage-I:} Pre-training backbone Latent VDM with autoencoder based on WebVid data \cite{BainNVZ21}.

\item \textbf{Stage-II:} Further pre-training the backbone VDM for text-conditioned video generation, based on WebVid data. 
We update the backbone diffusion model of Dysen-VDM, where the 3D-UNet includes an RGTrm encoder, and they all will be updated. 
There we will use the DSG annotations generated from Dysen.

\item \textbf{Stage-III:} Updating the overall Dysen-VDM with the dynamic scene managing (Dysen).

\end{compactitem}

\section{Experiments}
\label{Experiments}

\subsection{Setups}

\vspace{-1mm}
We experiment on two popular T2V datasets, including the UCF-101 \cite{abs-1212-0402} and MSR-VTT \cite{XuMYR16}.
In UCF-101, the given texts are the simple action labels.
In MSR-VTT, there are integral video caption sentences as input prompts.
To evaluate the action-complex scenario, we also adopt the ActivityNet data \cite{KrishnaHRFN17}, where each video connects to the descriptions with multiple actions (at least 3 actions), and the average text length is 50.4.
To relieve the computation burden, during the sampling phase in diffusion, we evenly sample 16 keyframes from a two-second clip, and then interpolate them twice with higher frame rates.
We perform image resizing and center cropping with a spatial resolution of 256$\times$256 for each input text.
The latent space is 32$\times$32$\times$4. 
The denoising sampling step $T$ is 1000.
In default, we use the ChatGPT (\textit{GPT-3.5} turbo) via OpenAI API.\footnote{\url{https://platform.openai.com}}
For action planning and scene imagination, we sample $D$=5 in-context demonstrations.
RGTrm takes $L$=12 layers and $k$=8 attention heads.
All dimensions are set as 768. 
Initial $\beta$ is set 0.5, and then decays gradually.

Following previous works \cite{abs-2304-08477,abs-2304-08818,he2022latent}, we use the Inception Score (IS) and Fr\'echet Video Distance (FVD) for UCF-101, and Fr\'echet Image Distance (FID) and CLIP similarity (CLIPSIM) for MSR-VTT.
We also use human evaluation for a more intuitive assessment of video quality.
We consider two types of settings: 1) zero-shot, where our pre-trained model makes predictions without tuning on on-demand training data;
2) directly fine-tuned on training data without large pre-training.
We consider several existing strong-performing T2V systems as our baselines, which are shown later.
Also, we re-implement several open-sourced baselines for further customized evaluations, including CogVideo \cite{abs-2205-15868}, VDM \cite{HoSGC0F22} and Latent-VDM \cite{RombachBLEO22}.
Scores from our implementations are averaged in five runs with random seeds, and the results of other baselines are copied from the raw papers.
All our training is conducted on 16 NVIDIA A100 GPUs.

\begin{table}[!t]
\vspace{-1mm}
\caption{
Zero-shot results on UCF-101 and MSR-VTT data.
The results of baselines are copied from their raw paper.
The best scores are marked in bold.
}
\vspace{-2mm}
\fontsize{9}{11}\selectfont
\setlength{\tabcolsep}{1.5mm}
\label{tab:Zero-shot}
\centering
\begin{tabular}{lcccc}
\hline
\multicolumn{1}{c}{\multirow{2}{*}{\bf Method}}&   \multicolumn{2}{c}{\bf UCF-101} & \multicolumn{2}{c}{\bf MSR-VTT} \\
\cmidrule(r){2-3}\cmidrule(r){4-5}
 & \bf IS ($\uparrow$) & \bf	FVD ($\downarrow$) & \bf FID ($\downarrow$) & \bf CLIPSIM ($\uparrow$)   \\
\hline
CogVideo \cite{abs-2205-15868}& 	25.27 & 	701.59 & 	23.59 & 	0.2631 \\   
MagicVideo \cite{abs-2211-11018} & 	/ & 	699.00 & 	/ & 	/ \\   
MakeVideo \cite{abs-2209-14792} & 	33.00 & 	367.23 & 	13.17 & 	0.3049 \\   
AlignLatent \cite{abs-2304-08818} & 	33.45 & 	550.61 & 	/ & 	0.2929 \\   
Latent-VDM \cite{RombachBLEO22} & 	/ & 	/ & 	14.25 & 	0.2756 \\ 
Latent-Shift \cite{abs-2304-08477} & 	/ & 	/ & 	15.23 & 	0.2773 \\   
VideoFactory \cite{wang2023videofactory} & 	/ &  410.00 & 	/ & 	0.3005 \\ 
InternVid \cite{wang2023internvid}  & 		21.04 & 616.51  & / & 	0.2951 \\ 
\cdashline{1-5}
\bf Dysen-VDM & \bf 35.57 &  \bf 325.42 & \bf 12.64 &  \bf 0.3204 \\   
\hline
\end{tabular}
\vspace{-1mm}
\end{table}

\begin{table}[!t]
\caption{
Fine-tuning results on UCF-101 without pre-taining.}
\vspace{-2mm}
\fontsize{9}{11}\selectfont
\setlength{\tabcolsep}{5.8mm}
\label{tab:Fine-tuned}
\centering
\begin{tabular}{lcc}
\hline
\bf Method & \bf IS ($\uparrow$) & \bf	FVD ($\downarrow$) \\
\hline
VideoGPT \cite{abs-2104-10157} & 	24.69 & 	/ \\   
TGANv2 \cite{SaitoSKK20} & 	26.60 & 	/ \\   
DIGAN \cite{YuTMKK0S22} & 	32.70 & 	577$\pm$22 \\   
MoCoGAN-HD \cite{TianRCO0MT21} & 	33.95 & 	700$\pm$24 \\   
VDM \cite{HoSGC0F22}	 & 57.80 & 	/ \\   
LVDM \cite{he2022latent} & 	27.00	 & 372$\pm$11 \\   
TATS \cite{GeHYYPJHP22} & 	79.28 & 	278$\pm$11 \\   
PVDM \cite{abs-2302-07685} & 	74.40 & 	343.60 \\   
ED-T2V \cite{liu2023ed} & 	83.36 & 	320.00 \\  
VideoGen \cite{li2023videogen} & 	82.78 & 	345.00\\  
Latent-VDM \cite{RombachBLEO22} & 	90.74 & 	358.34 \\   
Latent-Shift \cite{abs-2304-08477} & 	92.72 & 	360.04 \\   
\cdashline{1-3}
\bf Dysen-VDM & \bf 95.23 &  \bf 255.42 \\   
\hline
\end{tabular}
\vspace{-4mm}
\end{table}

\subsection{Main Comparisons and Observations}

\paragraph{Zero-shot Performance.}
We first present the comparison results on the zero-shot setting on UCF-101 and MSR-VTT datasets, respectively.
As shown in Table \ref{tab:Zero-shot}, Dysen-VDM outperforms the baselines on both IS and FVD metrics with big margins on UCF-101 data, where the given texts are the simple action labels, and the dynamic scene imagination capability is especially needed.
This shows the capability of our model.
Note that on MSR-VTT data we calculate the frame-level metrics between the testing captions and video frames, and we see that our system secures the best results.

\vspace{-3mm}
\paragraph{On-demand Fine-tuning Results.}
Table \ref{tab:Fine-tuned} further presents the results of the fine-tuned setting on UCF-101 data.
We see that with the on-demand training annotations, the winning scores of our system over the baselines become more clear.
In particular, Dysen-VDM model achieves 95.23 IS and 255.42 FVD scores, respectively, becoming a new state-of-the-art.

\begin{figure}[!t]
\centering
\includegraphics[width=0.98\columnwidth]{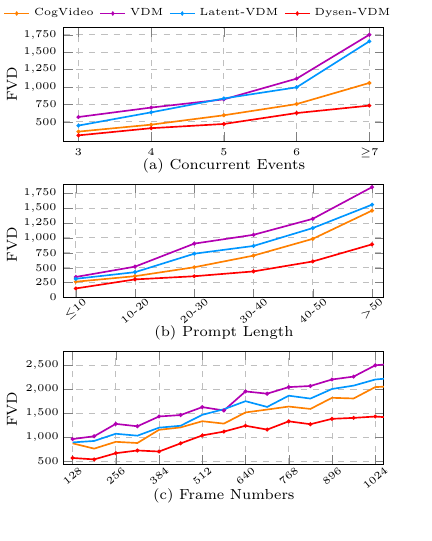}
\vspace{-1mm}
\caption{
Performance on the action-complex scene video generation of ActivityNet data.
}
\label{complex-action}
\vspace{-5mm}
\end{figure}

\vspace{-1mm}
\subsection{Results on Action-complex T2V Generation}

\vspace{-1mm}
Now we consider a more strict comparing setting of action-complex scenario.
We use the ActivityNet data under the fine-tuning setup.
We consider three different testing T2V scenarios:
1) the input texts containing multiple concurrent (or partially overlapped) actions,
2) the prompts having different lengths,\footnote{
As the text length in ActivityNet is no less than 20, we randomly add some test sets from MSR-VTT data.
}
and 3) generating different lengths of video frames.
We make comparisons with CogVideo, VDM and Latent-VDM, where the last two are diffusion-based T2V methods.
As plotted in Figure \ref{complex-action}, overall Dysen-VDM evidently shows stronger capability than the baseline methods, on all three tests of action-complex T2V generation.
We also see that the superiority becomes more clear when the cases go harder, i.e., with more co-occurred events, longer input prompts and longer video generation.
We note that CogVideo uses large pre-training, thus keeping comparatively better performance than the other two T2V diffusion models. 
In contrast, without additional pre-training, our system is enhanced with scene dynamics modeling
can still outperform CogVideo significantly.

\subsection{Human Evaluation}
\label{Human Evaluation}

\vspace{-1mm}
The standard automatic metrics could largely fail to fully assess the performance with respect to the temporal dynamics of generated videos.
We further show the human evaluation results on the ActivityNet test set, in terms of action faithfulness, scene richness, and movement fluency, which correspond to the issues shown in Figure \ref{intro}.
We ask ten people who have been trained with rating guidelines, to rate a generated video from 0-10 scales, and we average the final scores into 100 scales.
As seen in Table \ref{tab:Human}, overall, our system shows very exceptional capability on the complex-scene T2V generation than other comparing systems.
In particular, Dysen-VDM receives a high 92.4 score on the scene richness, surpassing CogVideo by 17.4, and also wins over Latent-VDM on action failthfulness by 15.9.
We can give the credit to the action planning and scene imagination mechanism in Dysen module.

\newcommand{\specialcell}[2][c]{%
  \begin{tabular}[#1]{@{}c@{}}#2\end{tabular}}

\begin{table}[!t]
\vspace{-2mm}
\caption{
Human evaluation on ActivityNet data.}
\vspace{-2mm}
\fontsize{9}{11}\selectfont
\setlength{\tabcolsep}{2.0mm}
\label{tab:Human}
\centering
\begin{tabular}{lccc}
\hline
& \bf \specialcell{Action\\Faithfulness}& \bf \specialcell{Scene\\Richness}& \bf \specialcell{Movement\\Fluency} \\
\hline
CogVideo \cite{abs-2205-15868} & 	67.5 & 	75.0	 & 81.5 \\   
VDM \cite{HoSGC0F22} & 	62.4 & 	58.8 & 	46.8 \\   
Latent-VDM \cite{RombachBLEO22} & 	70.7 & 	66.7 & 	60.1 \\   
\cdashline{1-4}
\bf Dysen-VDM & \bf 86.6 & \bf 92.4 & \bf 87.3 \\    
\hline
\end{tabular}
\vspace{-1mm}
\end{table}

\begin{table}[!t]
\caption{
Model ablation (fine-tuned results in FVD).
`w/o Dysen': degrading our system into the Latent-VDM model.
}
\vspace{-1.5mm}
\fontsize{9}{11}\selectfont
\setlength{\tabcolsep}{3mm}
\label{tab:Ablation}
\centering
\begin{tabular}{lll}
\hline
\multicolumn{1}{c}{\bf Item} & \multicolumn{1}{c}{\bf UCF-101} & \multicolumn{1}{c}{\bf ActivityNet} \\
\hline
\bf Dysen-VDM & \bf 255.42 &  \bf 485.48 \\  
\cdashline{1-3}
\quad w/o Dysen & 	346.40{\tiny(+90.98)} & 	627.30{\tiny(+141.82)} \\  
\quad w/o Scene Imagin. & 	332.92{\tiny(+77.50)} & 	597.83{\tiny(+112.35)} \\  
\qquad w/o SWC & 	292.16{\tiny(+36.74)} & 	533.22{\tiny(+47.74)} \\  
\quad w/o RL-based ICL & 	319.01{\tiny(+63.59)} & 	520.76{\tiny(+35.28)} \\  
\cdashline{1-3}
\quad RGTrm$\to$RGNN \cite{NicolicioiuDL19} & 	299.44{\tiny(+44.02)} & 	564.16{\tiny(+78.68)} \\  

\hline
\end{tabular}
\vspace{-2mm}
\end{table}

\begin{figure}[!t]
\centering
\includegraphics[width=0.99\columnwidth]{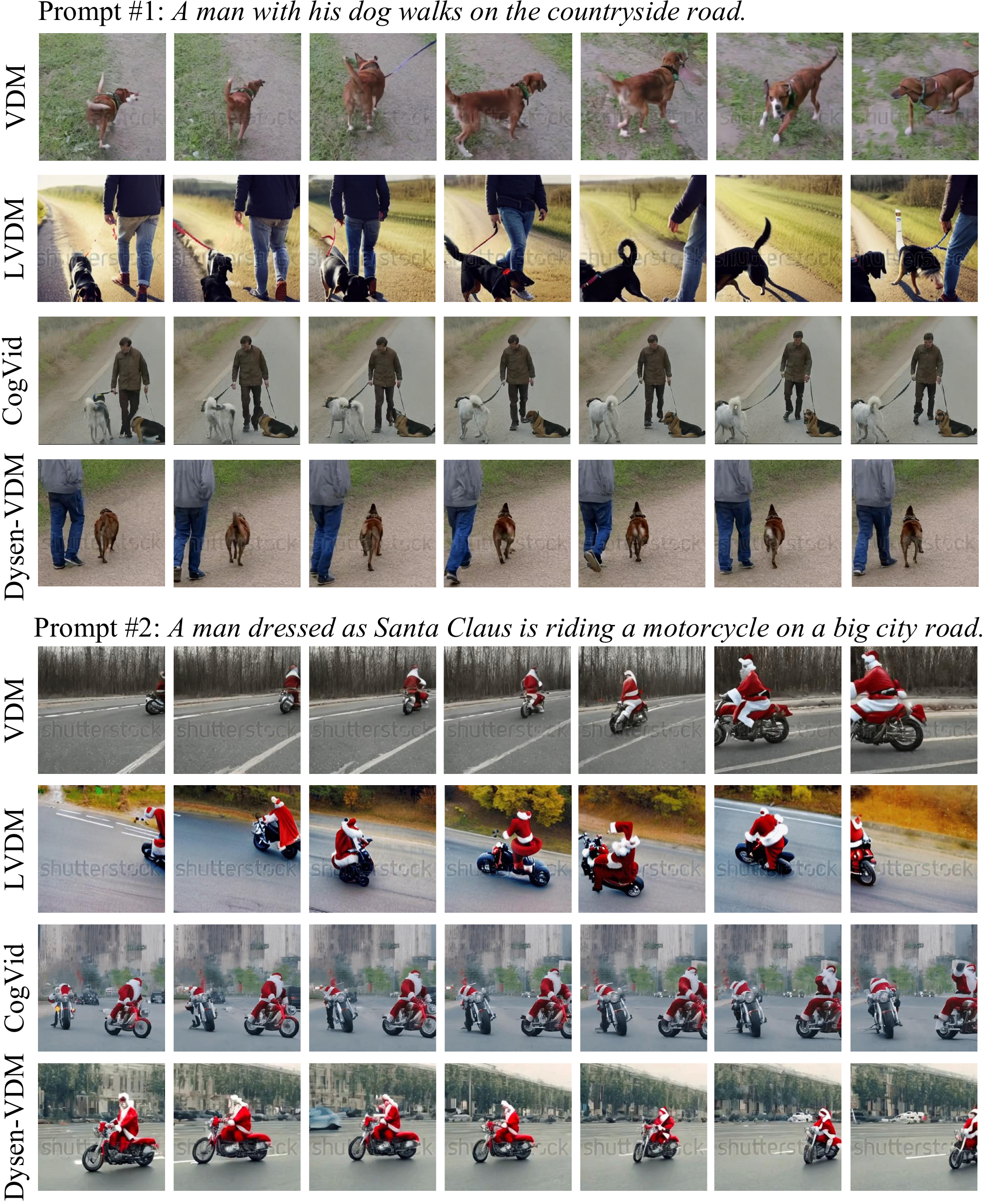}
\vspace{-1mm}
\caption{
Qualitative results on video generation with two pieces of examples.
Visit the live demos at {\url{http://haofei.vip/Dysen-VDM/}} for more cases.
}
\label{Case}
\vspace{-2mm}
\end{figure}

\subsection{System Ablations}

\vspace{-1mm}
We further conduct ablation studies to quantify the specific contribution of each design of our system.
As shown in Table \ref{tab:Ablation}, we can find that, first of all, removing the whole Dynsen module (then equal to the Latent-VDM model) results in the most crucial performance loss, with +90.98 FVD on UCF-101 and +141.82 FVD on ActivityNet.
This evidently verifies the efficacy of the Dynsen module and indirectly indicates that the core of high-quality T2V synthesis lies in modeling the motion dynamics. 
Further, removing step 3 of Dynsen, the scene imagination part, we see there are also significant drops, only second to the whole Dynsen.
When only without the sliding window context (SWC) mechanism, the performance can be also hurt, indicating the importance of generating reasonable and fine scene details for T2V.
Then, if canceling the RL optimization for ICL, the performance is hurt, especially on UCF-101 data, as the short labels require much more high-quality demonstrations to prompt ChatGPT for correct action planning and scene enrichment.
Finally, the proposed RGTrm also serves irreplaceable roles for the fine-grained spatio-temporal feature encoding.

\vspace{-1mm}
\subsection{Qualitative Results}

\vspace{-1mm}
To gain a more direct understanding of how better our system succeeds in generating videos with smooth and complex movements, we present qualitative comparisons with the baseline models in Figure \ref{Case}.
As can be observed, Dysen-VLM has exhibited overall better performance.
For both two prompts, our model shows smooth video frames with accurate motions occurring in order,
while the videos by LVDM have quite jumpy transitions between different frames, and also the dynamic video scenes are not delicate, with some erroneous actions.
The main reason largely lies in whether the T2V system models the intricate temporal dynamics.
Also we see that the video by baseline may fail to faithfully reflect all the predicates mentioned in the input instructions; baseline missed certain actions.
For example in prompt \#1, `\emph{man walks}' is missed by VDM.

\begin{figure}[!t]
\centering
\includegraphics[width=0.9\columnwidth]{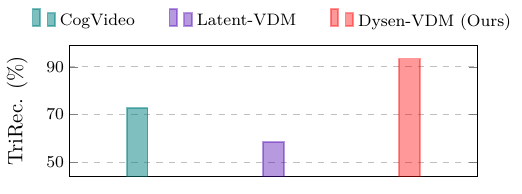}
\vspace{-1mm}
\caption{
Aligning recall rate (TriRec.) of `\emph{sub.-prdc.-obj.}' structures between prompt and generated video frames.
}
\label{SG-controllable}
\vspace{-1mm}
\end{figure}

\begin{figure}[!t]
\centering
\includegraphics[width=0.82\columnwidth]{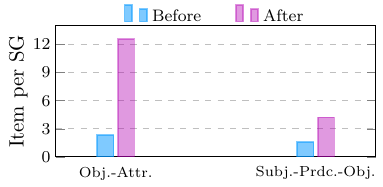}
\vspace{-1mm}
\caption{
The number of SG structures (`\emph{obj.-attr.}' and `\emph{sub.-prdc-obj.}') before and after scene imagination.
}
\label{changed-scene}
\vspace{-2mm}
\end{figure}

\subsection{In-depth Analyses}
\label{In-depth Analyses}

\paragraph{Controllability with DSG.}
SG has shown to have better semantic controllability, due to its semantically structured representations \cite{JohnsonKSLSBL15,YoonKJLHPK21,abs-2211-11138}.
Here we examine such superiority of our system where our dynamic scene enhancement is built based on DSG.
Following \cite{wu2024imagine}, we use the \emph{Triplet Recall} (TriRec.) to measure the fine-grained `\emph{subject-predicate-object}' structure recall rate between the SGs of input texts and video frames.
Given a set of ground truth triplets, denoted $G^{GT}$, and TriRec. is computed as:
\begin{equation}
 \text{TriRec.} = \frac{|G^{PT}\cap G^{GT}|}{|G^{GT}|} \,,
\end{equation}
where $G^{PT}$ are the relation triplets of the SG in the generated video DSG by a visual SG parser.
As plotted in Figure \ref{SG-controllable}, Dysen-VDM achieves the highest score than two baselines with clear margins.

\vspace{-3mm}
\paragraph{Change of Scenes.} 
We then make statistics of the structure changes before and after the scene imagination in Dysen module.
We mainly observe the `\emph{object-attribute}' and `\emph{subject-predicate-object}' SG structures, where the former reflects the static contents, and the latter reflects the dynamic scenes.
From Figure \ref{changed-scene} we learn that both two types of SG structures are increased in numbers via scene imagination by LLM.
This indicates a clear scene enrichment, leading to better video generation.

\begin{figure}[!t]
\centering
\includegraphics[width=0.82\columnwidth]{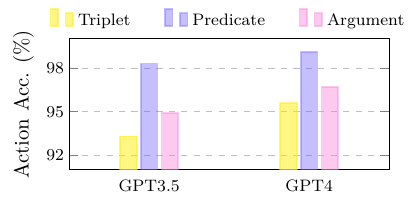}
\vspace{-1mm}
\caption{
Accuracy of action annotation (overall triplet, predicate and argument (subject\&object) using GPT3.5 and GPT4.
}
\label{action-acc}
\vspace{-2mm}
\end{figure}

\vspace{-3mm}
\paragraph{Action Parsing via ChatGPT.}
Action planning, as the first step, is pivotal to the overall following performance, where we employ the ChatGPT for inducing majorly-occurring events/actions.
Here we analyze the quality of action parsing, and the influence of using GPT3.5 and GPT4.
We randomly select 100 samples from the MSR-VTT data, and then compare between the ChatGPT-generated annotations and manually annotated ones.
From Figure \ref{action-acc} we see that both GPT3.5 and GPT4 shows quite satisfied accurate induction, with GPT4 advancing more slightly.

\section{Conclusion}

\vspace{-1mm}
In this work, we enhance the intricate temporal dynamics modeling of video diffusion models (VDMs) for text-to-video (T2V) synthesis.
Inspired by human intuition of video filming, we design an innovative dynamic scene manager (Dysen) module, which performs three steps of temporal dynamics understanding:
first extracting key actions with proper time-order arrangement;
second, transforming the ordered actions into dynamic scene graph (DSG) representations;
third, enriching the DSG scenes with sufficient reasonable details.
We implement the Dysen based on ChatGPT, for human-level temporal dynamics understanding, where the in-context learning is optimized via reinforcement learning.
Finally, we newly devise a recurrent graph Transformer to learn the fine-grained delicate spatio-temporal features from DSG, and then integrate them into the backbone T2V DM for video generation.
Experiments on three T2V datasets show that our dynamics-aware video DM achieves new best results, especially performs stronger in scenarios with complex actions.

\section{Acknowledgments}

\vspace{-1mm}
This research is supported by CCF-Baidu Open Fund and CCF-Baichuan Yingbo Innovation Research Funding.

\newpage

{
\bibliographystyle{ieee_fullname}
\bibliography{refs}
}

\clearpage

\begin{figure*}[!t]
\centering
\includegraphics[width=0.97\textwidth]{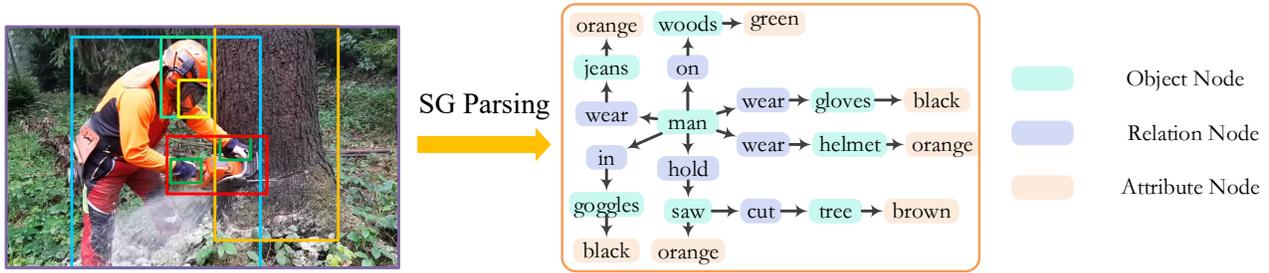}
\caption{
Illustration of a visual SG.
}
\label{app-sg}
\end{figure*}

\appendix

\section{Extended Technical Specification}

In this part, we provide some extended technical details of the proposed methods.

\subsection{Specification on T2V Latent Diffusion Model}
\label{Specification on Text-to-video Latent Diffusion Model}

First, we can formalize the T2V task as generating an video $X$=$\{x_1,\cdots, x_F\} \in \mathbb{R}^{F \times H \times W \times C}$ given the input prompt text $Y$=$\{w_1,\cdots,w_S\}$.
Here $F, H, W, C$ are the frame length, height, width and channel number of the video respectively.
We use the existing latent diffusion model (LDM) to accomplish the task.
As shown in Figure \ref{framework}, LDM consists of a diffusion (forward) process and a denoising (reverse)
process in the video latent space.
An encoder $\mathcal{E}$ maps the video frames into the lower-dimension latent space, i.e., $Z_0 = \{\mathcal{E}(X)\}$, and later a decoder $\mathcal{D}$ re-maps the latent variable to the video, $X = \{\mathcal{D}(Z_0)\}$.

\paragraph{Diffusion Process.}
The diffusion process transforms the input video into noise.
Given the compressed latent code $Z_0$, LDM gradually corrupts it into a pure Gaussian noise over $T$ steps by increasingly adding noisy, i.e., Markov chain.
The noised latent variable at step $t \sim [1, T]$ can be written as:
\begin{equation}
 Z_t = \sqrt{\hat{\alpha}_t} x + \sqrt{1-\hat{\alpha}_t} \epsilon_t \,,
\end{equation}
with 
\begin{equation}
 \hat{\alpha}_t = \sum_{k=1}^t \alpha_t \,, \quad \epsilon_t \sim \mathcal{N}(0, I) \,,
\end{equation}
where $\alpha_t \in (0,1)$ is a corresponding diffusion coefficiency.
The diffusion process can be simplified as: $q(Z_{1:T} |Z_0) = \prod_{t=1}^{T}q(Z_t|Z_{t-1})$.

\paragraph{Denoising Process.} 
The denoising process then restores the noise into the video reversely.
The learned reverse process $ p_{\theta}(Z_{0:T}) = p(Z_T)\prod_{t=1}^{T} p_{\theta} (Z_{t-1}|Z_t, Y)$ gradually reduces the noise towards the data distribution conditioned on the text $Y$.
Our T2V LDM is trained on video-text pairs ($X, Y$) to gradually estimate the noise $\epsilon$ added to the latent code given a noisy latent $Z_t$, timestep $t$, and conditioning text $Y$:
\begin{equation}
 \mathcal{L}_{\text{\scriptsize LDM}} = \mathbb{E}_{Z \sim \mathcal{E}(X),Y,\epsilon,t} \left[ \, || \epsilon -\epsilon_{\theta} (Z_t, t, \mathcal{C}(Y))   ||^2 \, \right]  \,,
\end{equation}
where $\mathcal{C}(Y)$ denotes a text encoder that models the conditional text, and the denoising network $\varepsilon_\theta (\cdot)$ is often implemented via a 3D-UNet.


The original 3D-UNet \cite{HoSGC0F22} has a spatial-temporal feature modeling.
The practical implementation is to use a 3D convolution along the spatial dimension, and followed by a temporal attention.
Specifically, given a set of $F$ video frames, we apply a shared 3D convolution for all the frames to extract the spatial features. 
After that, we assign a set of distribution adjustment parameters to adjust the mean and variance for the intermediate features of every single frame via:
\begin{equation}
 Z^i_t = \bm{W}^i \text{Conv3D}(  Z^i_t ) + b^i  \,,
\end{equation}
where the convolutions are performed over 3D patches $Z^i_t$.
Then, with a set of given video frame features, $Z_t \in \mathbb{R}^{F\times H\times W\times C}$, we apply the temporal attention to the spatial location across different frames to model their dynamics.
Specifically, we first reshape  $Z_t$ into shape of $HW \times \#Heads \times F \times \frac{C}{\#Heads}$.
We then obtain their query $Q_t$, key $K_t$, and value $V_t$ embeddings used in the self-attention via three linear transformations. 
We calculate the temporal attention matrix $H_t$ via:
\begin{equation}
    H_t = \text{Softmax}(\frac{Q_t \cdot K_t} {\sqrt{d}}) \cdot M \,, 
\end{equation}
where $M$ is an lower triangular matric with $M_{i,j}=0$ if $i>j$ else 1.
With the implementation of the mask, the present token is only affected by the previous tokens and independent from the future tokens since the frames are arranged based on their temporal sequence.

\begin{figure*}[!t]
\centering
\includegraphics[width=1\textwidth]{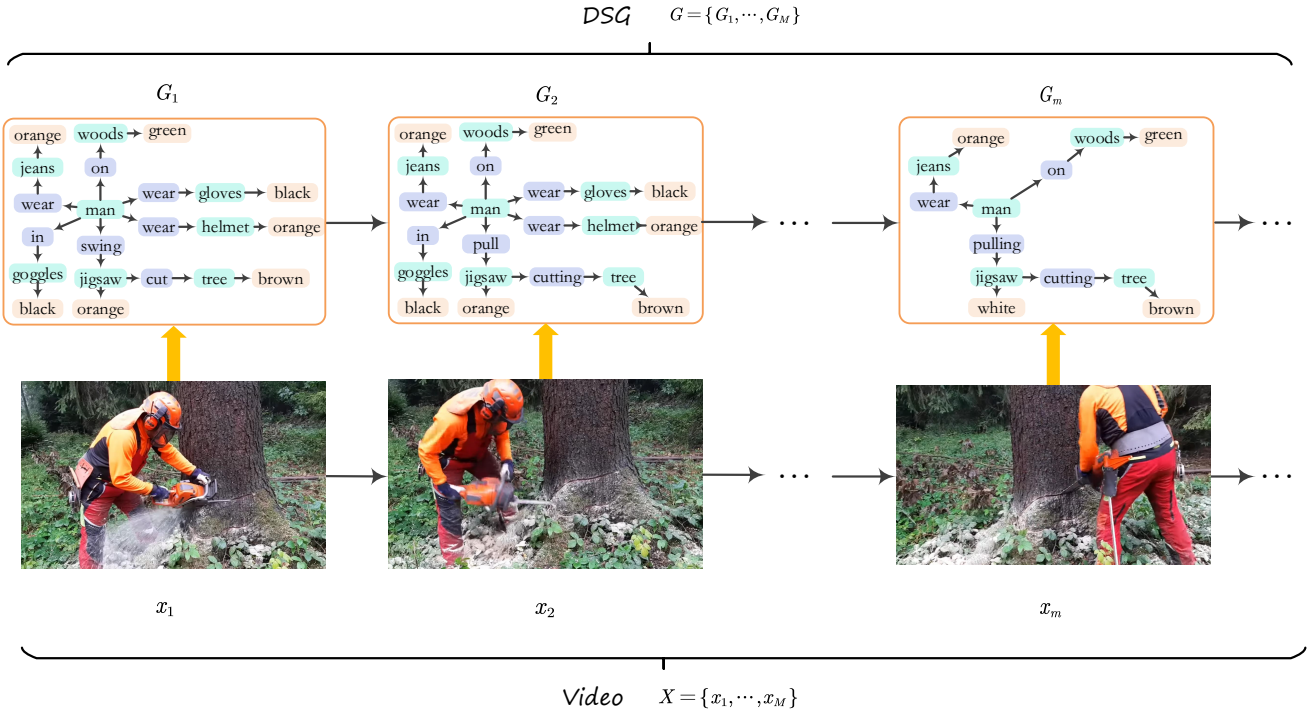}
\caption{
DSG is a list of single visual SG in the temporal order.
}
\label{app-dsg}
\end{figure*}

\subsection{Specification on Dynamic Scene Graph Representation}
\label{Specification on Dynamic Scene Graph Representation}

DSG \cite{JiK0N20} is a list of single visual SG \cite{JohnsonGF18} of each video frame, organized in time-sequential order.
We can denote an DSG as $G$=$\{G_1,\cdots,G_M\}$, with each single SG ($G_m$) corresponding to the frame ($x_m$).
An SG contains three types of nodes, i.e., \emph{object}, \emph{attribute}, and \emph{relation}, in which some scene objects are connected in certain relations, forming the spatially semantic triplets `\emph{subject-predicate-object}'.
Also, objects are directly linked with the attribute nodes as the modifiers.
Figure \ref{app-sg} illustrates the visual SG.

Since a video comes with inherent continuity of actions, the SG structure in DSG is always temporal-consistent across frames.
This characterizes DSGs with spatial\&temporal modeling.
Figure \ref{app-dsg} visualizes a DSG of a video.

\subsection{Prompting ChatGPT with In-context Learning}
\label{Prompting ChatGPT with In-context Learning}

In $\S$\ref{Dynamic Scene Manager} we introduce the Dysen module for action planning, event-to-DSG conversion, and scene enrichment, during which we use the in-context learning (ICL) to elicit knowledge from ChatGPT.
Here we elaborate further on the specific designs, including 1) the ICL for action planning, and 2) two ICLs for scene enrichment, containing the step-wise scene imagination and global scene polishment.

For each task, we write the prompts, including 
1) a job description (\texttt{Instruction}), 
2) a few input-output in-context examples (\texttt{Demonstration}), 
and 3) the desired testing instance (\texttt{Test}).
By feeding the ICL prompts into ChatGPT, we expect to obtain the desired outputs in the demonstrated format.
Note that we pre-explored many other different job instruction prompts, and the current (shown in this paper) version of instructions helps best elicit task outputs.
Also, for each ICL, we select five examples as demonstrations, which can be enough to prompt the ChatGPT.

\begin{figure*}[!h]
\centering
\includegraphics[width=0.98\textwidth]{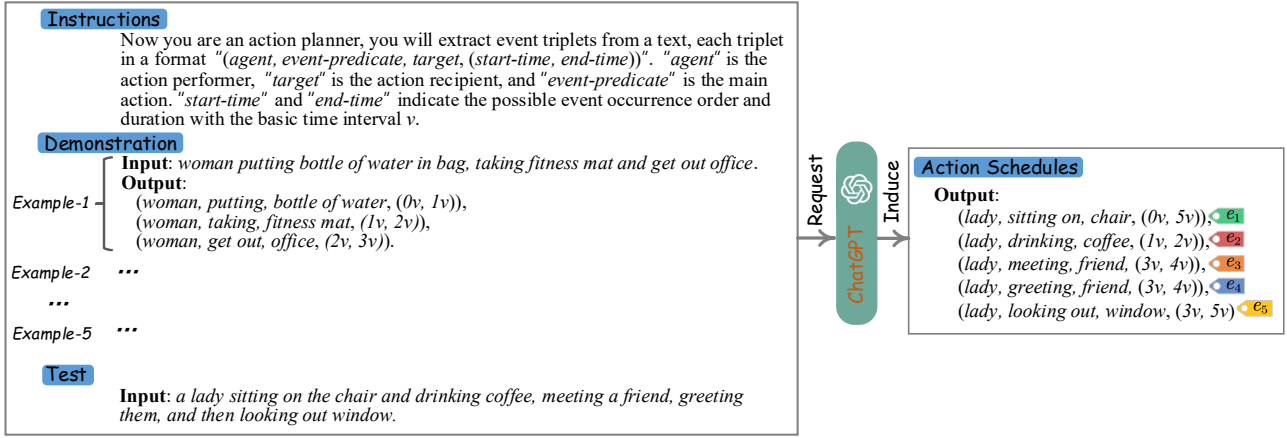}
\vspace{-1.5mm}
\caption{
Illustration of the ICL for action planning.
}
\label{app-ICL-A}
\end{figure*}

\paragraph{ICL Design for Action Planning.}

In Figure \ref{app-ICL-A} we show the complete ICL illustration for action planning.
Note that the demonstration examples should come with the `gold' annotations, so as to correctly guide the ChatGPT to induce high-quality output.
In our implementation, for the action planning, we first obtain the majorly-occurred events (i.e., actions) from the input texts via ChatGPT.
Note that this can be a very easy task within the natural language processing area, and we simply treat this as the gold annotations.
Then, we use an off-the-shelf best-performing video moment localization model \cite{zhang2021multi} to parse the video starting and ending positions for each event expression.
Via this, we obtain the action planning annotations for the demonstrations.

\begin{figure*}[!h]
\centering
\includegraphics[width=0.98\textwidth]{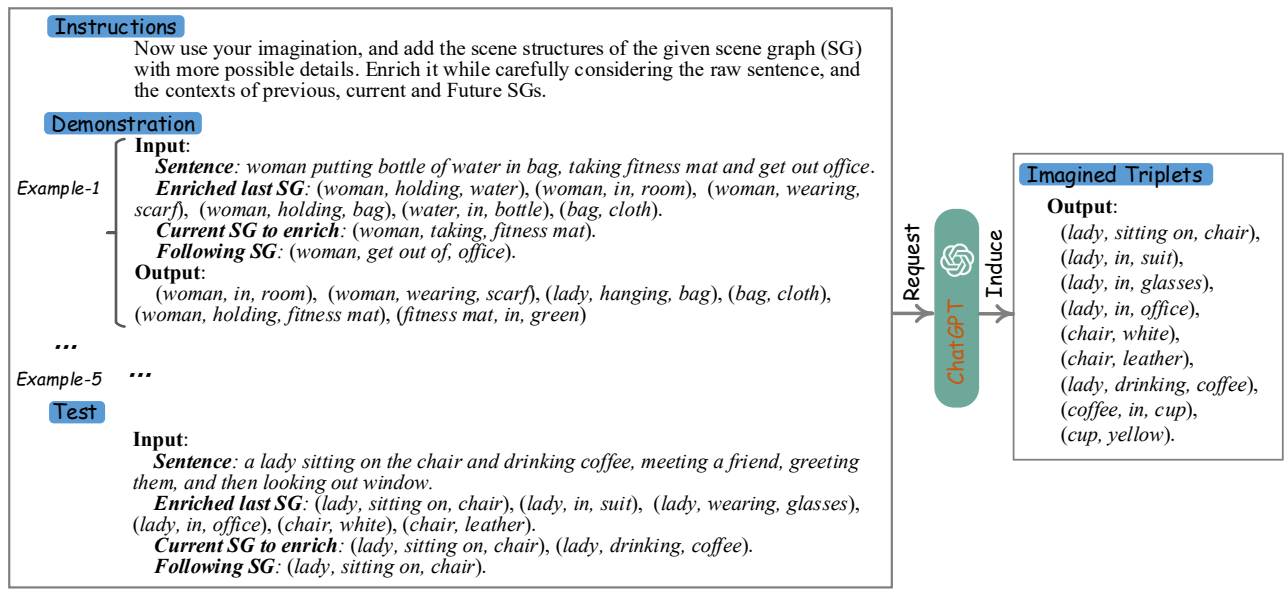}
\vspace{-1.5mm}
\caption{
Illustration of the ICL for step-wise scene imagination.
}
\label{app-ICL-B}
\end{figure*}

\paragraph{ICL Design for Step-wise Scene Imagination.}

In Figure \ref{app-ICL-B} we show the full illustration of the ICL for step-wise scene imagination.
We note that the output triplets after imagination are the full-scale triplets, including the raw ones of the unenriched SG.
This means, we will overwrite the raw SG with the output triplets, which cover both the \emph{add} and \emph{change} operations.

\begin{figure*}[!h]
\centering
\includegraphics[width=1\textwidth]{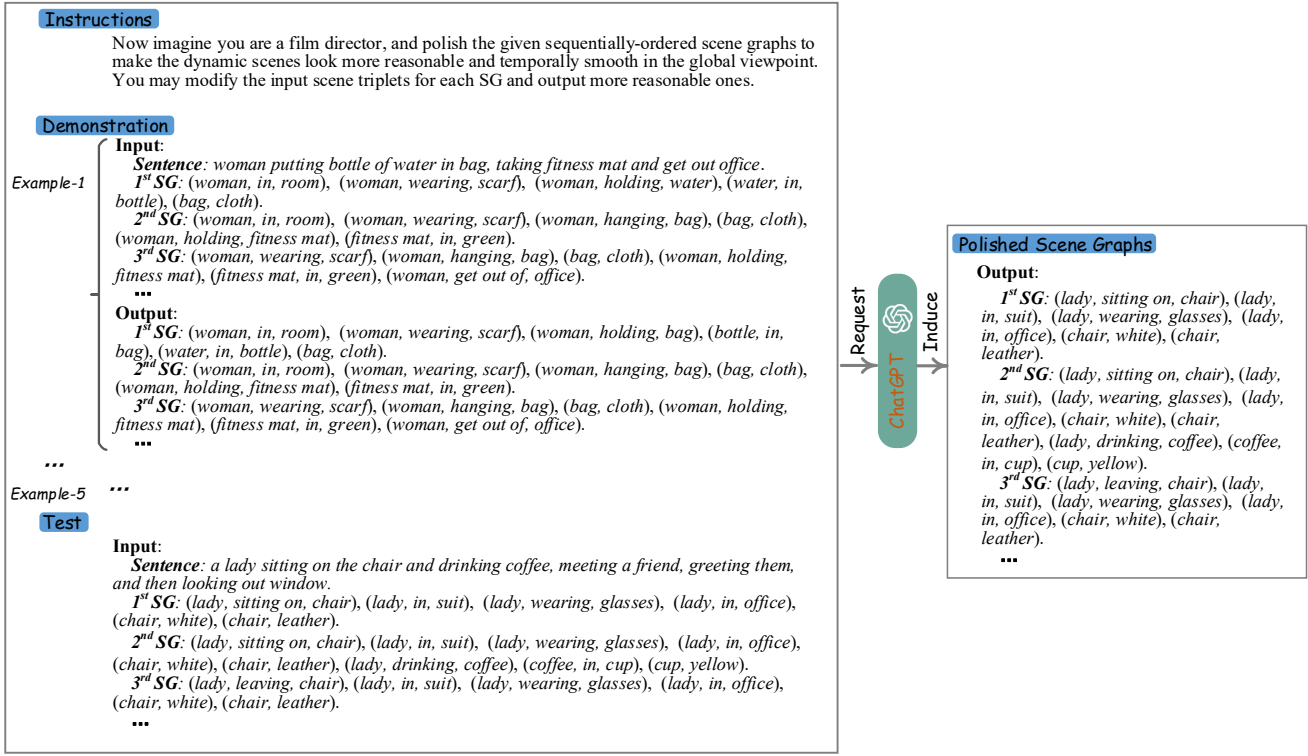}
\caption{
Illustration of the ICL for global scene polishment.
}
\label{app-ICL-C}
\vspace{-3mm}
\end{figure*}

\paragraph{ICL Design for Global Scene Polishment.}

In Figure \ref{app-ICL-C} we show the full illustration of the ICL for global scene polishment.
Same to the ICL for step-wise scene imagination, we also take the output DSG as the refined DSG.
The `gold' annotations for demonstrations of the scene imagination are constructed by parsing the video via external DSG parser \cite{JiK0N20}.
We employ the FasterRCNN \cite{RenHGS15} as an object detector to obtain all the object nodes, 
and use MOTIFS \cite{TangNHSZ20} as a relation classifier to obtain the relation labels (nodes) as well as the relational edges.
We then use an attribute classifier to obtain attribute nodes.
We filter the object, attribute, and relation annotations by keeping those that appear more than 500 times in the training set. 
This helps screen the less-informative noises.

\section{Experiment Specification}
\label{Experiment Specification}

In this part we extend the description of the experimental settings.

\subsection{Evaluation}

In our experiments, we used the following three types of evaluation metrics.

\paragraph{Automatic Metrics.}
Following previous work \cite{abs-2304-08477,abs-2304-08818,he2022latent}, we use the Inception Score (IS) and Fr\'echet Video Distance (FVD) for UCF-101, and Fr\'echet Image Distance (FID) and CLIP similarity (CLIPSIM) for MSR-VTT.

\begin{compactitem}
\item \textbf{IS} \cite{SalimansGZCRCC16} evaluates the distribution of the frame images of our generated videos.\footnote{\url{https://torchmetrics.readthedocs.io/en/stable/image/inception_score.html}} 
Following previous work on video synthesis, we used a C3D \cite{TranBFTP15} model trained on UCF-101 to calculate a video version of the inception score, which is calculated from 10k samples using the official code of TGANv2.\footnote{\url{ https://github.com/pfnet-research/tgan2}}

\item \textbf{FVD} measures the similarity between real and generated videos \cite{abs-1812-01717}. 
For the generated videos (16 frames at 30 FPS), we extract features from a pre-trained I3D action classification model.\footnote{\url{https://www.dropbox.com/s/ge9e5ujwgetktms/i3d_torchscript.pt?dl=1’}}

\item \textbf{FID} \cite{HeuselRUNH17} measures the Fr\'echet Distance between the distribution of the frames between synthetic and gold videos in the feature space of a pre-trained Inception v3 network. 
Practically, we employ \emph{pytorch-fid}\footnote{\url{https://github.com/mseitzer/pytorch-fid}} to calculate the FID score.

\item \textbf{CLIPSIM} \cite{HesselHFBC21} is also used for the quantitative analysis of the semantic correctness of the text-to-video generation on MSR-VTT data.
We take into account the reference-free scores via CLIP \cite{RadfordKHRGASAM21}. In this paper, we use the officially released code\footnote{\url{https://github.com/jmhessel/clipscore}} to calculate the CLIP score.
We generate 2,990 videos (16 frames at 30 FPS) by using one random prompt per example. 
We then average the CLIPSIM score of the 47,840 frames. 
We use the ViT-B/32 \cite{RadfordKHRGASAM21} model to compute the CLIP score.

\end{compactitem}

\paragraph{Human Evaluation Criterion.}
In \S\ref{Human Evaluation} we also adopt the human evaluation, i.e., user study, for more intuitive assessments of video quality.
On the ActivityNet test set, we compare our model with baseline systems.
We randomly select 50 text-video pairs (videos containing both the gold-standard ones and the generated ones), and ask ten participants (native English speakers) who have been trained with rating guidelines, to rate a generated video.
Specifically, we design a Likert 10-scale metric to measure the target aspect: 
1-Can't be worse,
2-Terrible,
3-Poor, 
4-Little poor, 
5-Average,
6-Better than average,
7-Adequate, 
8-Good, 
9-Very good,
10-Excellent. 
For each result, we take the average.
We mainly measure the quality of videos in terms of \emph{action faithfulness}, \emph{scene richness} and \emph{movement fluency}, each of which is defined as:
\begin{compactitem}
    \item \textbf{Action faithfulness}: 
    Do the visual actions played in the video coincide with the raw instruction of the input texts? Is there any point missed or incorrectly generated?
    
    \item \textbf{Scene richness}: 
    Are the visual scenes rich? Are there vivid and enough background or foreground details in the video frames?
    
    \item \textbf{Movement fluency}: 
    Are the video dynamics of actions fluent? Is the video footage smooth? Are the behaviors presented in a continuous and seamless manner?
    
\end{compactitem}

\paragraph{Triplet Recall Rate.}
In Figure \ref{SG-controllable} we use the \emph{Triplet Recall} (TriRec.) to measure the fine-grained `\emph{subject-predicate-object}' structure recall rate between the SGs of input texts and video frames.
Technically, TriRec. measures the percentage of the correct relation triplet among all the relation triplets between two given SGs. 
Given a set of ground truth triplets (\emph{subject-relation-object}), denoted $G^{GT}$, and the TriRec. is computed as:
\begin{equation}
 \text{TriRec.} = \frac{|G^{PT}\cap G^{GT}|}{|G^{GT}|} \,,
\end{equation}
where $G^{PT}$ are the relation triplets of the SG in the generated video DSG by a visual SG parser.

\subsection{Baseline Specification}
\label{Baseline specification}

We mainly compare with the currently strong-performing T2V systems as our baselines, which are divided into two groups: non-diffusion-based T2V, and diffusion-based T2V.

\begin{compactitem}
    
    \item \textbf{Non-diffusion-based T2V Methods}
    \begin{compactitem}
        \item \textbf{VideoGPT} \cite{abs-2104-10157} is a two-stage model: it encodes videos as a sequence of discrete latent vectors using VQ-VAE and learns the autoregressive model with these sequences via Transformer.
        
        \item \textbf{TGANv2} \cite{SaitoSKK20} is a computation-efficient video GAN based on designing submodules for a generator and a discriminator.
        
        \item \textbf{DIGAN} \cite{YuTMKK0S22} is a video GAN which exploits the concept of implicit neural representations and computation-efficient discriminators.

        \item \textbf{MoCoGAN-HD} \cite{TianRCO0MT21} uses a strong image generator for high-resolution image synthesis. 
        The model generates videos by modeling trajectories in the latent space of the generator.

        \item \textbf{TATS} \cite{GeHYYPJHP22} is a new VQGAN for videos and trains an autoregressive model to learn the latent distribution.

        \item \textbf{CogVideo} \cite{abs-2205-15868} is a large-scale pre-trained text-to-video generative model based on Transformer with dual-channel attention.

        \item \textbf{InternVid}  \cite{wang2023internvid} is a video-text representation learning model based on ViT-L via contrastive learning.

    \end{compactitem}

    \vspace{5pt}
    \item \textbf{Diffusion-based T2V Methods}
    \begin{compactitem}
        \item \textbf{VDM} \cite{HoSGC0F22} extends the image diffusion models for video generation by integrating a 3D-UNet architecture based on 3D convolutional layers.

        \item \textbf{LVDM} \cite{he2022latent} is built upon the latent diffusion models with a hierarchical diffusion process in the latent space for generating longer videos.
        
        \item \textbf{MakeVideo} \cite{abs-2209-14792} directly translates the tremendous recent progress in Text-to-Image (T2I) generation to T2V without training T2V from scratch.

        \item \textbf{MagicVideo} \cite{abs-2211-11018} is a latent diffusion based T2V model, which takes 2D convolution + adaptor block operation and a directed self-attention module for the spatial-temopral learning.
        
        \item \textbf{AlignLatent} \cite{abs-2304-08818} is also a latent diffusion based T2V model, which leverages the pre-trained image DMs for video generators by inserting temporal layers that learn to align images in a temporally consistent manner.

        \item \textbf{ED-T2V}  \cite{liu2023ed} is an efficient training framework for diffusion-based T2V generation, which is built on a pretrained text-to-image generation model. 
        
        \item \textbf{VideoGen}  \cite{li2023videogen} is a cascaded latent diffusion module conditioned on both the reference image and the text prompt, for generating latent video representations, followed by a flow-based temporal upsampling step to improve the temporal resolution.
        
        \item \textbf{VideoFactory}  \cite{wang2023videofactory}         strengthens the interaction between spatial and temporal perceptions by utilizing a swapped cross-attention mechanism in 3D windows that alternates the "query" role between spatial and temporal blocks, enabling mutual reinforcement for each other.

        \item \textbf{Latent-VDM}: we implement a T2V baseline of latent video diffusion model based on the latent diffusion \cite{RombachBLEO22}, with the widely-adopted spatial convolution and temporal attention.
        
        \item \textbf{Latent-Shift} \cite{abs-2304-08477} adds a parameter-free temporal shift module onto the existing latent video diffusion model to enhance the motion dynamics learning of video generation.
        
    \end{compactitem}
    
\end{compactitem}

To enable further customized evaluations and experiments, we also re-implement some open-sourced baselines, including CogVideo\footnote{\url{https://github.com/THUDM/CogVideo}} \cite{abs-2205-15868}, VDM\footnote{\url{https://github.com/lucidrains/video-diffusion-pytorch}} \cite{HoSGC0F22} and Latent-VDM\footnote{\url{https://github.com/nateraw/stable-diffusion-videos}} \cite{RombachBLEO22}.

\section{Extended Experiments}

\subsection{Visualization of DSG-guided Controllable Video Synthesis}
\label{More Qualitative Results}

\begin{sidewaysfigure*}
\centering
\includegraphics[width=1\textwidth]{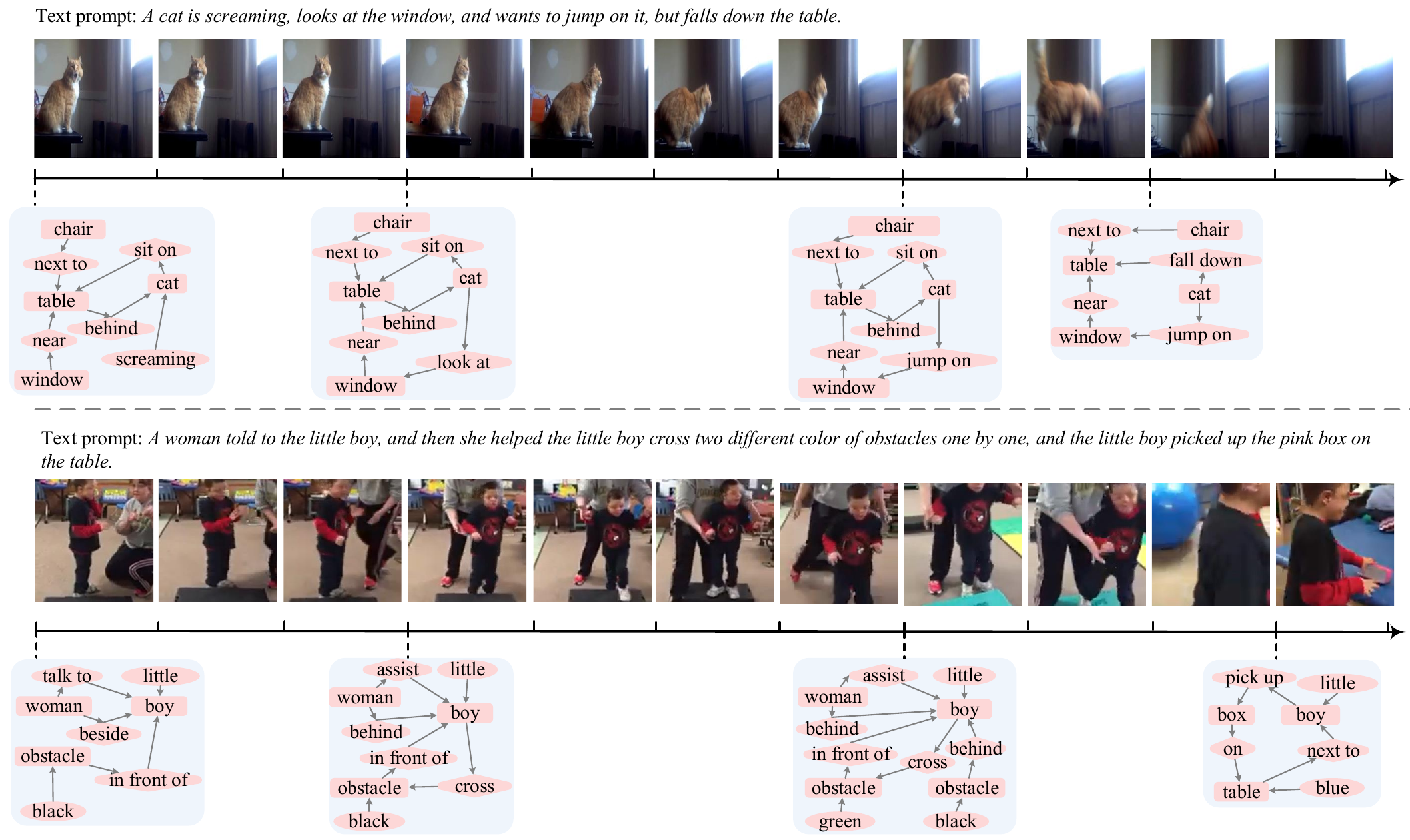}
\caption{
Visualization of generated videos with the induced and enriched DSG structures.
}
\label{dsg_process}
\end{sidewaysfigure*}

In Figure \ref{dsg_process} we show the video frames generated by our Dysen-VDM, along with which we visualize the DSGs induced and enriched by the Dysen module.

\subsection{Failure Analysis}
\label{Error Analysis}

While Dysen-VDM helps achieve overall improved performance in most of cases, it will err sometimes.
Here we summarize the typical failure cases of Dysen-VDM that were made during our experiments.

\begin{compactitem}

\item \textbf{Type-1:} Due to the limitations of LLM, sometimes it may hallucinate, leading to errors in scene understanding. 
The imagined DSG quality is relatively low, which, in turn, affects the quality of the generated video.

\item \textbf{Type-2:} DSG is very proficient at generating realistic videos. 
However, there are some abstract scenes, such as cartoon-style videos, that cannot be supported by the structured triplet representations of SG. DSG struggles to effectively improve specific artistic styles in certain frames.

\end{compactitem}

But fortunately, in most of the T2V scenarios, Dysen-VDM can advance.
The integration of structured DSG representations with rich details (from LLMs' imagination) empowers the system with highly controllable content generation and high-quality video dynamics.

\end{document}